# Copy the dynamics using a learning machine

Hong Zhao

Department of Physics, Xiamen University, Xiamen 361005, China

zhaoh@xmu.edu.cn

ABSTRACT

**Is it possible to generally construct a dynamical system to simulate a black system without recovering the equations of motion of the latter? Here we show that this goal can be approached by a learning machine. Trained by a set of input-output responses or a segment of time series of a black system, a learning machine can be served as a copy system to mimic the dynamics of various black systems. It can not only behave as the black system at the parameter set that the training data are made, but also recur the evolution history of the black system. As a result, the learning machine provides an effective way for prediction, and enables one to probe the global dynamics of a black system. These findings have significance for practical systems whose equations of motion cannot be approached accurately. Examples of copying the dynamics of an artificial neural network, the Lorenz system, and a variable star are given. Our idea paves a possible way towards copy a living brain.**

Hearing the shape of a drum is a focus topic of the well-studied inverse problem [1,2]. The key point is whether the shape of the drumhead can be recovered uniquely by the sound it makes. This question has been clarified, i.e., different shapes may make identical sounds [2-4], and thus it is not possible to generally recover the equations of motion of a black system based on observed data without sufficient *a prior* knowledge. The finding of different systems can produce the identical output, on the other hand, inspires a more interested problem: Whether one can generally construct another system to simulate the dynamics of a black system without recovering its equations of motion? With such a system, one may predict the output of the black system, and even may probe the global dynamics of the black system. More attractively, it may bring the science-fiction story of copying a living human brain using a machine into a topic of serious research, i.e., to recover the dynamics of a biological system using a robot 'brain'.

Here we show that properly constructed learning machines can fulfill the task. Learning machines have wide applications in various fields [5-13]. For the purpose of copying the dynamics, we present a general framework of learning machines and introduce a flexible algorithm to train a learning machine using a set of input-output (I-O) responses or a segment of time series of the black system. As examples we first show the application to copy an artificial neural network which has a memory of 60000 handwriting digits [10,14]. Just trained by a set of random I-O responses from the artificial neural network, a learning machine can recovery its memory. The second application is to recovery the celebrated Lorenz system [15-16]. Besides the recovering of the dynamics with high accuracy at the parameter set that the training data are made, we exhibit the recurrence of its historical dynamics in the course of training process, implying that the learning machine can even replay the evolution history of the black system. We finally report the result of mimicking the evolution dynamics of the light curve of a chaotic variable star [17-20], and reveal the possible evolution modes of a class of variable stars in their history. Variable star is popular in cosmos. Its information can be used to deduce even more fundamental knowledge about our universe in general.

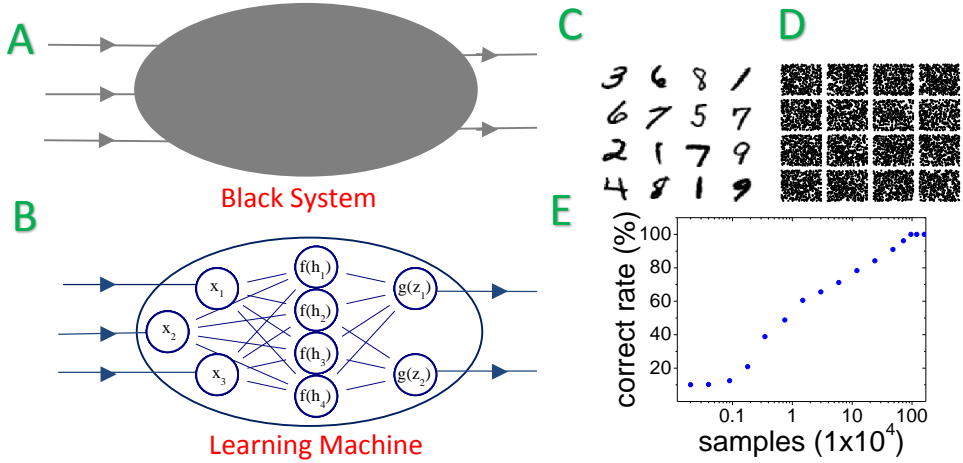

Fig. 1. Copying an artificial neural network. A: Black system, B: Learning machine. C: Examples of memories of the black system. D: Examples of training patterns. E: The recovering rate of memories as a function of training sample amount.

The architecture of the learning machine is illustrated in Fig. 1B. It is an $M-N-L$ three-layer network, with $M$, $N$, and $L$ neurons in the input, hidden, and output layers respectively. The variables $x_i(t), h_j(t), z_k(t)$ ($i=1,\ldots,M$; $j=1,\ldots,N$; $k=1,\ldots,L$) are the inputs of neurons in the three layers correspondingly, the function $f$ is the neuron transfer function of hidden-layer neurons, and $g$ represents the operator to out-layer neurons. By setting $g(z_k) = z_k(t+1)$ and $g(z_k) = dz_k(t)/dt$, one obtains an iterative learning machine (ILM) $\vec{z}(t+1) = \Phi(\Lambda, \vec{x}(t))$ and a differential learning machine (DLM) $d\vec{z}(t)/dt = \Phi(\Lambda, \vec{x}(t))$, respectively, where $\Lambda$ represents the parameter set. They can be turned into self-evolving systems by setting $\vec{z}(t) = \vec{x}(t)$ with $M=L$. These two kinds of machines are applicable for black systems with I-O responses, $(\vec{x}^\mu, \vec{z}^\mu), \mu=1,\ldots,P$, being measurable. When only a segment of time series is measurable from the black system, we apply delay coordinate $\vec{x}(t) = (x(t), x(t-\tau),\ldots, x(t-M\tau))$ to be the input vector to construct a $M$-$N$-1 delay-differential learning machine (DDLM) $dz(t)/dt = \Phi(\Lambda, x(t), x(t-\tau), x(t-M\tau))$, where $\tau$ is the inter-record gap length. It can also be turned into self-evolving systems by setting $z(t) = x(t)$.

The training is to find a parameter set $\Lambda$ which guarantees the learning machine responses identically as the black system does at the training set, while gains the best generalization ability to response also correctly to general inputs as the black system does. For this purpose, we have developed a training method (21, method and supplementary materials). It applies a Monte Carlo algorithm to train the machine response to the training set, and applies the so-called design risk minimization principle to gain the generalization ability. The control parameters can fulfill such a goal is called as the optimal control parameter set. The most remarkable difference from the conventional methods is that we can construct sufficient large learning machines without over training, which enables us to find the copy system with high freedom (arc, SI).

**Copy a neural network**. The black system is a 784-1000-10 forward iterative neural network for recognizing the handwriting digits in the MNIST data base [14]. The hidden-layer neuron transfer function is the sigmoid one $f(h) = \tanh(h)$. The original samples are represented by 28×28 black-white patterns, some of them are shown in figure 1C. 10 neurons in the output layer identify the 10 numbers (0-9) correspondingly by checking which one has the maximum output. We train the black system to correctly memory the 60,000 MNIST samples. We then input random 784-dimensional binary vectors (Figure 1D as examples) to the black system to measure the corresponding outputs. These I-O responses are applied to train a 784-3000-10 ILM. The neuron transfer function in the hidden layer is chosen to be $f(h) = \exp(-h^2)$, which is different from that of the black system. Figure 1E shows the recognition rate of the ILM to the MNIST samples as a function of the amount of training samples. We see that all of the handwriting digits that the black system has memorized are correctly recovered with the increase of training samples. Note that the sample amount we applied is negligible comparing to the input-vector space with $2^{784}$ states. Thus the dynamics of a neural network can be recovered by a learning machine with different structure and inner function. The same treatment can be applied to recovery the dynamics of **attractor neural networks (SI).**

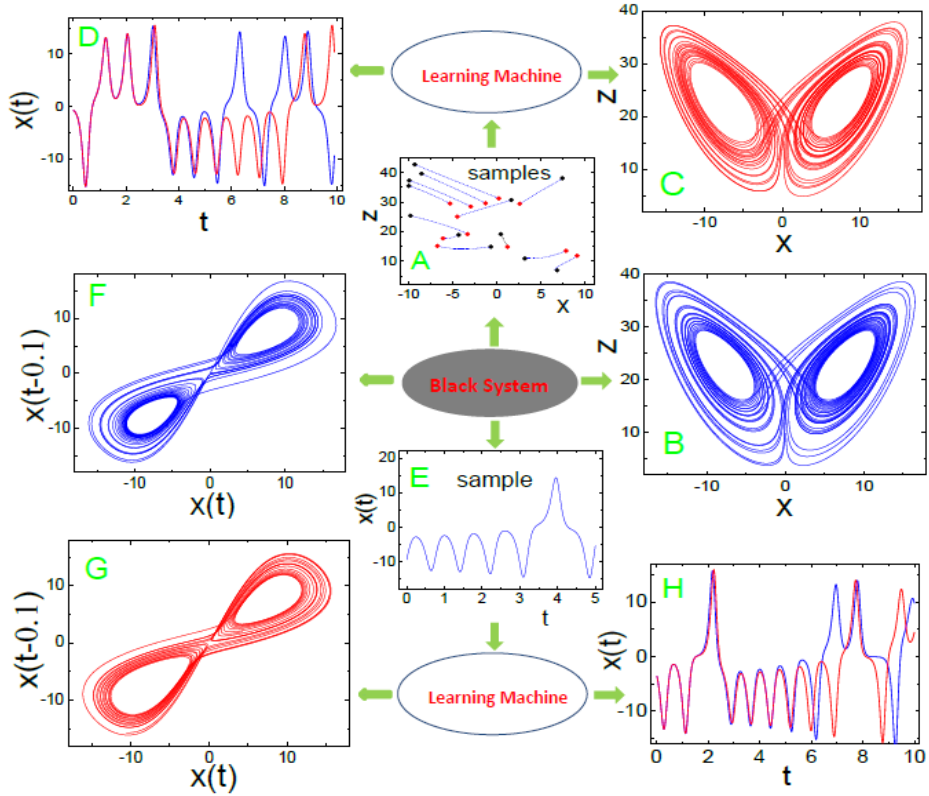

Figure 2. Copying the Lorenz system. A: Training samples for the DLM. This plot shows the projections of several training samples on the *x-z* plane. B and C: The attractors of the black system and the DLM projected to *x-z* plane. D: Predicted evolution curve (red) by the DLM comparing to that (blue) of the black system. E: The training sample for the DDLM. F and G: The attractors of the black system and the DDLM represented by delay coordinates. H: Predicted evolution curve (red) by the DDLM comparing to that (blue) of the black system.

**Copy the Lorenz system.** The Lorenz system is a prototype model of chaotic motion. It is described by equations, $dx/dt = -\sigma(x-y)$, $dy/dt = -xz + Rx - y$ and $dz/dt = xy - bz$. When the I-O responses are measurable, we evolve the system over an interval of $t=0.05$ from an initial point $(x(0), y(0), z(0))$ to obtain an output point $(x(t), y(t), z(t))$ to get a training sample. Initial points are chosen randomly from intervals $x(0), y(0) \in (-20, 20), z(0) \in (0, 40)$. At the standard chaotic parameter set $(\sigma, R, b) = (10, 28, 8/3)$, we prepare 100 such samples (see Figure 2A for examples) to train a 3-3000-3 DLM with $f(h) = \exp(-h^2)$. The DLM is evolved using the standard Runge-Kuttan integration program. Figure 2B-2C show the strange attractors of the Lorenz equations and that of the learning machine, respectively. By visual inspection, one cannot distinguish which one is the original attractor.

In the case only a segment of time series of a variable can be measured, we apply an $M-3000-1$ DDLM to recover the dynamics. The sample is a time serious of length $t=5$ (Figure 2E) recorded with $\tau=0.001$. With $M=100$, we obtain 4900 samples. The learning machine is trained to correctly response these samples. The trajectory is simulated by the Runge-Kuttan integration program for delay-differential equations. After the training goal is approached, we apply the last sample to be initial condition to evolve the learning machine. Figure 2F-2G show the attractors represented by the delayed coordinate $[x(t), x(t-0.1)]$ for the black system and the DDLM, respectively. The two attractors show no qualitative difference. This example indicates that a differential system can be simulated by a delay-differential system, i.e., the learning machine and the black system need not to be governed by the same type of dynamical equations. Note that the training set is very small, far from being ergodic to the configuration space.

Applying the learning machine, one can predict the trajectory of the black system. In Figure 2(D) we plot the evolution curves of $x(t)$ started from the same point for the Lorenz system (blue) and the DLM (red) respectively. We see that the two curves do not separate till $t>6.5$. This result is corresponding to using the Lorenz equations per se with the same σ and $b$ but a slight different $R$ ( $R=28+0.02$ ). Similarly, applying a segment of length $t=0.01$ from the record to be the initial condition, the DDLM can track the trajectory over a period of $t=5.5$, as Figure 2H shows.

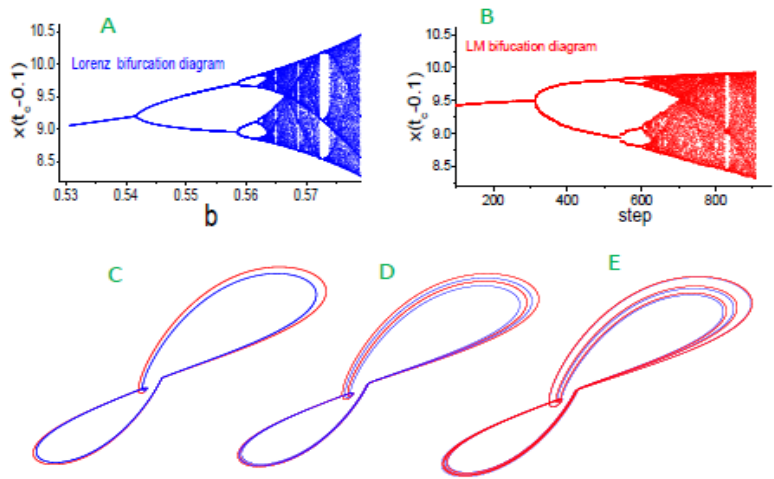

Figure 3. Recurring the evolution history of the black system. A: A segment of bifurcation diagram of the Lorenz system along parameter b. B: The bifurcation diagram of the DDLM trained by a segment of time series at b=0.58. The horizontal axis is the training time (measured by MC steps). C: Period one, D: period two, and E: period three solutions of the Lorenz system (red) and learning machine (blue), correspondingly. The trajectories are represented by $x(t-0.1)-x(t)$.

The learning machine may reveal more information about the black system, i.e., though trained by the data made at a specific parameter set, it may replay the evolution history of the black system. In more detail, before the training target is approached, the learning machine can already mimic the dynamics of the black system close the parameter set that the data are made. To show this phenomenon, in Fig. 3A we plot a segment of the bifurcation diagram of the Lorenz system along the axis of the system parameter $b$ with another tow parameters fixed at $(\sigma, R) = (10, 28)$, where $x(t_c - 0.1)$ is the delayed coordinate of the section point that the $x$ variable across the section of $x=5$ at $t_c$. We then apply a segment of $t=30$ of the time serious measured at $b=0.58$ (the end point of Figure 3A) to train DDLM, and plot a series of $x(t_c - 0.1)$ after a period of training. Figure 3B plots the bifurcation diagram of the DDLM as a function of the training time. We see that Figure 3A is recovered qualitatively. Furthermore, as examples, Figure 3C-3E show the period-1, period-2 and period-3 attractors of the Lorenz system and the DDLM picking up alone the bifurcation diagram. We find that the corresponding attractors are indeed qualitatively consistent. This fact indicates that before the dynamics inducing the current training data is approached, the learning machine has already turn to be a dynamical system belong to the same class of the black system.

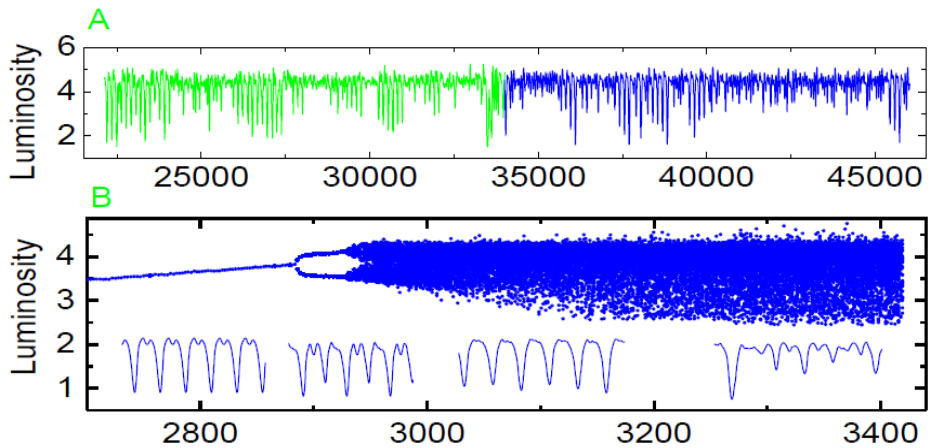

Figure 4. Copying a variable star. A: The green line is the original record of the light curve of R Scuti variable star. The blue line is the out put of the learning machine trained by the recorded data. B: The bifurcation diagram of the learning machine as a function of the training time. The insets are period-one, period-two, quasi-period and chaotic solutions along the bifurcation diagram.

**Copy the Variable Star**. The R Scuti is a chaotic variable star having records of the light curve over 100 years, a segment of which from 1960 is shown in Figure 4A by the green curve. The curve is smoothed by 10 days records. Using this segment to train a 100﹣ 300‑1 DILM with $f(h) = \tanh(h)$, the predicted subsequent evolution is also shown in Figure 4A by the blue curve. We see that the leaning machine can do mimic the dynamics of the variable star.

Figure 4B shows the bifurcation diagram of the learning machine along the axis of training time. We find the period-one solution (the first inset) and its period-doubling bifurcation (the second inset). The system evaluates to chaos by Hopf-like bifurcation, i.e., after just the period-doubling bifurcation it appears the quasi-periodic like motion, as the third inset shows. In the chaotic region, the learning machine may show a variety of evolution patterns, the last inset shows such an example. Supposing that this bifurcation diagram do represent in some extent the evolution history, one can conclude that this variable star has experienced periodic motion and its period-doubling bifurcation, as well as quasi-periodic motions before showing present chaotic light curve. This guess cannot be verified directly due to the limit of records. However, it is reasonable in view of the fact that these kind of light curves appear in different variable stars, which can be interpreted as that they are undergoing different stages of evolution. Recovering the dynamics from a time serious has more important practical applications, such as copy the sun's dynamics using the sunspot history record to predict its feature evolution (SI).

In conclusion, copying the dynamics of various black systems using a sufficient large learning machine is possible. It provides the best way for prediction of the consequence motion, as like using the black system itself. It provides also a way to probe the evolution history of a black system by using the present data.

In the past decade, several human brain projects have been launched [22-23]. With the development of these projects, detail knowledge about the brain will be revealed. Whatever detail, however, exactly rebuilding an organic system with $10^{10}$ neurons is impossible, let alone to recovery the memory. Our studies pave a way to copy a brain with a learning machine. Even based on current technology, we can expect to recover living neural systems of some simplest low-form life, since their neural networks have been explored in detail, and the state-of-the-art technologies have made accurately measurement of I-O response of a single neuron within the experimental reach.